# HILGEN: Hierarchically-Informed Data Generation for Biomedical NER Using Knowledgebases and Large Language Models

Yao Ge, PhD[1], Yuting Guo, MS[1], Sudeshna Das, PhD[1], Swati Rajwal, MTech[2], Selen Bozkurt, PhD[1], Abeed Sarker, PhD[1,3]

[1]Department of Biomedical Informatics, School of Medicine, Emory University, Atlanta, GA; [2]Department of Biomedical Informatics, School of Medicine, Emory University, Atlanta, GA; [3]Department of Biomedical Engineering, Georgia Institute of Technology and Emory University, Atlanta, GA

**Abstract**

*We present HILGEN, a Hierarchically-Informed Data Generation approach that combines domain knowledge from the Unified Medical Language System (UMLS) with synthetic data generated by large language models (LLMs), specifically GPT-3.5. Our approach leverages UMLS's hierarchical structure to expand training data with related concepts, while incorporating contextual information from LLMs through targeted prompts aimed at automatically generating synthetic examples for sparsely occurring named entities. The performance of the HILGEN approach was evaluated across four biomedical NER datasets (MIMIC III, BC5CDR, NCBI-Disease, and Med-Mentions) using BERT-Large and DANN (Data Augmentation with Nearest Neighbor Classifier) models, applying various data generation strategies, including UMLS, GPT-3.5, and their best ensemble. For the BERT-Large model, incorporating UMLS led to an average $F_1$ score improvement of 40.36%, while using GPT-3.5 resulted in a comparable average increase of 40.52%. The Best-Ensemble approach using BERT-Large achieved the highest improvement, with an average increase of 42.29%. DANN model's $F_1$ score improved by 22.74% on average using the UMLS-only approach. The GPT-3.5-based method resulted in a 21.53% increase, and the Best-Ensemble DANN model showed a more notable improvement, with an average increase of 25.03%. Our proposed HILGEN approach improves NER performance in few-shot settings without requiring additional manually annotated data. Our experiments demonstrate that an effective strategy for optimizing biomedical NER is to combine biomedical knowledge curated in the past, such as the UMLS, and generative LLMs to create synthetic training instances. Our future research will focus on exploring additional innovative synthetic data generation strategies for further improving NER performance.*

**Introduction**

Named Entity Recognition (NER) plays a critical role in extracting relevant entities from unstructured data, transforming raw clinical data into actionable insights. NER identifies and categorizes entities into predefined categories such as drug names, genes, adverse drug events (ADEs), symptoms, and reasons for drug prescriptions.[1,2] Few-shot learning techniques are particularly valuable in this context, enabling effective NER with limited annotated data. Few-shot learning in biomedical NER contrasts with traditional lexicon-based approaches, which may struggle with lexical variants or ambiguous expressions in larger datasets, and deep learning models that require large amounts of training data.[3,4] For instance, in medical diagnosis, few-shot learning has been used to develop models that can make accurate predictions with only a few examples,[5,6] which is especially beneficial for rare or emerging diseases.[7]

Recent advancements highlight the potential of large language models (LLMs) such as Generative Pre-trained Transformer (GPT), for few-shot NER. LLMs excel in generating natural language across domains and tasks, and their adaptability is enhanced by prompt-based strategies, which can significantly improve accuracy.[8] The ability of LLMs to generate coherent and contextually relevant text offers new opportunities to address the intricacies of NLP tasks involving medical data. By generating synthetic texts that closely mimic real-world medical text, LLMs can provide additional training data that enhances the performance of downstream tasks, such as NER and other critical applications in healthcare.[9] However, LLMs often face problems like hallucination[10] and homogenisation[11] when dealing with specialized medical concepts.

In this paper, we introduce HILGEN (Hierarchically-Informed Data Generation for Biomedical NER Using Knowledgebases and LLMs), a novel method that infuses domain knowledge and hierarchical information from the UMLS, with synthetic data generated by LLMs. We then train a supervised system to detect and evaluate the quality of the synthetic data generated using UMLS and LLMs, both separately and in ensemble approaches. Our approach aims to enrich the representations of sparsely occurring medical concepts, thereby enhancing performance in few-shot settings for biomedical NER tasks. Leveraging knowledge from both domain-specific knowledgebases and contextual information extracted by LLMs is relatively nascent. HILGEN aims to enhance the representation of rare medical concepts by integrating hierarchical domain knowledge with contextually relevant synthetic data. Thus, it addresses the data limitations in few-shot learning settings, significantly improving NER performance in biomedical texts. By incorporating UMLS knowledge with LLMs, the risk of hallucination can be effectively mitigated, as the LLMs are anchored to validated medical concepts, leading to more accurate and reliable outputs.

**Background and Significance**

*Few-shot Learning for Named Entity Recognition*

Few-shot learning, also referred to as low-shot learning, is a machine learning paradigm where models learn to make predictions on a new class with only a small number of examples,[6,12] unlike traditional deep learning, which requires extensive data. The goal of few-shot learning is to train a model that can generalize to new classes with only a few examples, which makes it particularly useful for many NER tasks within the medical domain, where obtaining large amounts of data for each class can be challenging, although it is conceptually possible.[5] Early research in few-shot learning within biomedical NLP faced challenges due to the complexities of natural language data containing domain-specific terminologies and associations.[4] Prior knowledge has been identified as crucial[13,14] to combat this issue. Initial approaches, such as meta-learning and metric learning, leveraged prior knowledge at different levels to generalize to new tasks with limited data.[14] Meta-learning[15] has been one of the most common framework for the early stage of few-shot learning research, which is trained using a set of training tasks. Other approaches such as matching networks,[16] which use embedding functions to generalize knowledge, prototypical networks,[12] which generate prototype representations of classes to address overfitting issues, and transfer learning were considered mainstream directions in the field of few-shot learning[17] prior to the extensive utilization of LLMs.

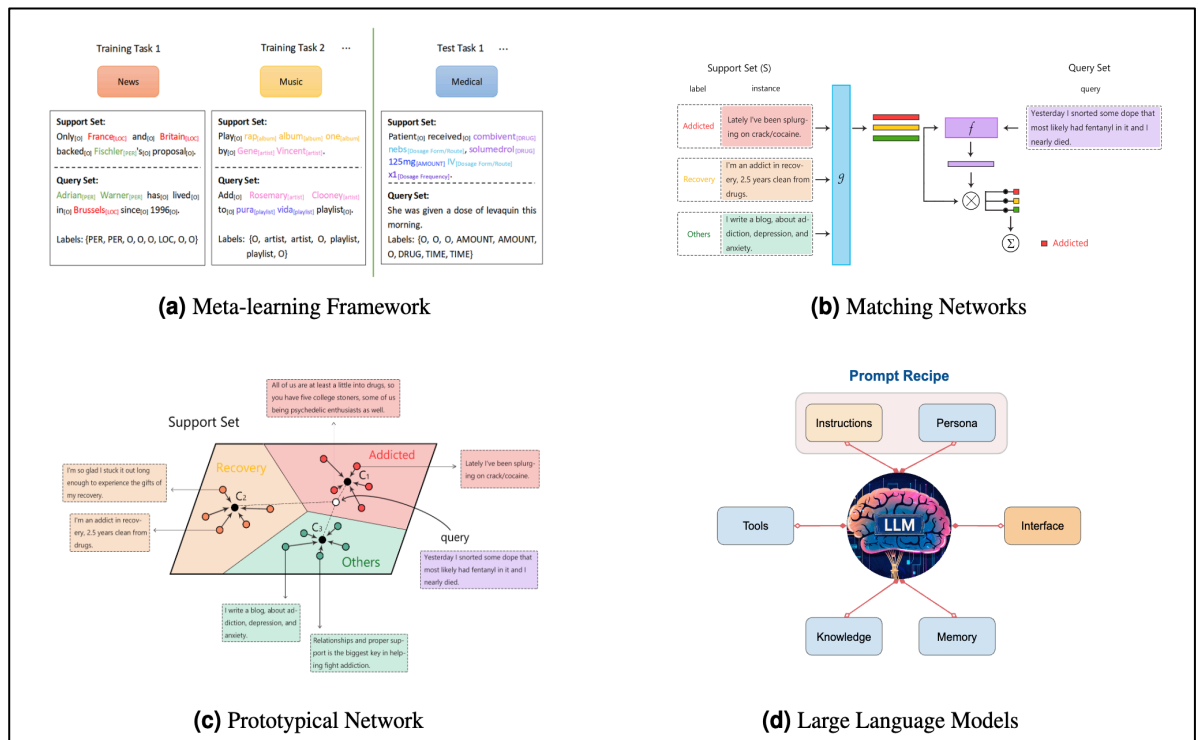

**Figure 1.** Shifts from three popular few-shot learning models to large language models. (a) Meta-learning: Each task mimics the few-shot scenario and can be completely non-overlapping. (b) Matching networks: A small support set contains some instances with their labels. Given a query, the goal is to calculate a value indicating whether the instance belongs to a given class. (c) Prototypical network: A class's prototype is the mean of its support set in the embedding space. (d) Typical Large Language Model structure: The prompt guides how the agent will proceed with the task and process the output.

The emergence of LLMs has largely shifted research focus towards prompt-based learning, which has shown promise in few-shot NLP.[18,19] Figure 1 shows the shifts from three popular few-shot learning models to LLMs. This approach relies on the ability of LLMs to generate text and provide contextually relevant responses with minimal training data for fine-tuning, offering a promising solution to the challenges posed by few-shot settings. The potential of LLMs and prompt-based strategies in few-shot settings is demonstrated by techniques like LM-BFF,[20] which fine-tunes models using prompts, and PPT,[21] which enhances prompt effectiveness through unsupervised pre-training. Incorporating biomedical knowledge bases like UMLS has also been explored,[22,23] showing improvements over general-purpose models. Integrating knowledge from both domain-specific knowledge bases and in-context information extracted by LLMs, however, is still relatively new. This presents a research gap that may have significant utility for challenging NLP tasks. We explore this potential utility on the task of few-shot NER using multiple biomedical datasets.

*UMLS in Biomedical Natural Language Processing*

UMLS is widely used in biomedical NLP for its comprehensive repository of medical terminologies, concepts, and relationships, serving as a critical resource for tasks like NER, information extraction, and text classification. UMLS integrates biomedical vocabularies and standards, providing a framework for disambiguation and unification of terminologies, which facilitates NLP systems in processing clinical and research data more accurately.[24] UMLS provides a structured repository containing over two million concepts, including synonyms, hierarchical relationships, and semantic types, making it an invaluable resource for disambiguating and standardizing medical terms. By leveraging UMLS in this paper, we ground the LLM-generated examples in accurate medical context, ensuring that the representations of medical entities remain semantically coherent and clinically relevant. This integration enables for the dynamic generation of enriched examples informed by UMLS, enhancing the model's understanding of rare medical terms and its ability to generalize across diverse biomedical datasets. Furthermore, the use of UMLS helps mitigate potential biases inherent in the training data of LLMs by providing a more balanced perspective on medical knowledge, ultimately resulting in more reliable outcomes in biomedical NER tasks.

*Large Language Models*

LLMs have driven substantial progress in the field of NLP, demonstrating exceptional performance across a wide range of tasks by generating and comprehending natural language at scale. Examples of this advancement include models like GPT-3,[25] GPT-4,[26] LaMDA,[27] and LLaMA[28] exemplify this advancement, which vary in architecture, data size, and training strategies, yet all benefit from the vast and diverse text data used during training. Nonetheless, LLMs encounter several notable challenges alongside their achievements. One prominent issue is hallucination, which occurs when LLMs generate content that is either nonsensical or unfaithful to the source material,[29] a problem observed in various applications. While this issue has been recognized and discussed within the research community, comprehensive studies and effective mitigation strategies are still under development.

This paper aims to contribute to this ongoing research by leveraging domain knowledge from UMLS with synthetic data generated by LLMs. Specifically, we propose to create synthetic datasets that can mimic the linguistic and semantic features of clinical texts, thereby offering a novel approach to data augmentation.

*Synthetic Data Generation*

Data augmentation and transfer learning are widely used techniques in machine learning to address data scarcity by generating or utilizing additional data, such as synthetic or noisy data, to improve data representation and diversity.[30,31] However, synthetic datasets often struggle to capture the naturalness and realism of human-written texts, particularly in clinical domains, and may introduce biases that affect the validity of downstream tasks. LLMs have been explored for generating clinical text, leveraging their capacity to store and produce health-related information. Recent studies have demonstrated the potential of LLMs in data augmentation for clinical tasks,[32] employing techniques such as the label-to-data method to mitigate the scarcity and sensitivity of medical data.

Traditional data augmentation approaches using pretrained language models often involve fine-tuning on existing datasets to generate synthetic data.[33,34] More recent methods focus on generating synthetic data with minimal supervision, using carefully crafted prompts or reverse tasks to produce high-quality data points.[35] Unlike these approaches, our work augments data by incorporating domain knowledge from UMLS alongside LLMs, leading to the generation of high-quality, task-relevant synthetic data.

**Materials and Methods**
The proposed approach, HILGEN (Hierarchically-Informed Data Generation for Biomedical NER Using Knowledge-bases and LLMs), infuses domain knowledge and hierarchical information from the UMLS,[36] along with synthetic data generated by LLMs. HILGEN is specialized for few-shot biomedical NER and aims to enrich the representation of sparsely occurring medical concepts by leveraging knowledge from both a human-curated knowledge base (UMLS) and an LLM.

*Approach*

The overall architecture of our HILGEN approach is shown in Figure 2. Prior to training, 5-shot examples are sampled from a given dataset, and relevant entities are extracted. Two distinct processing pipelines then generate synthetic examples based on the provided examples. UMLS-based data generation involves leveraging its hierarchical information and structured knowledge, and semantic network to automatically retrieve concepts related to named entities in the few-shot training data. GPT-based data generation leverages the generative capabilities of LLMs to produce contextually relevant synthetic linguistic structures from the few-shot examples. The new synthetic instances are then added to the original few-shot training data, and the models are fine-tuned on this augmented dataset. While UMLS-based data generation augments the data with domain-specific knowledge, GPT-based data generation allows us to leverage vast amounts of open-domain data. We provide further methodological details below.

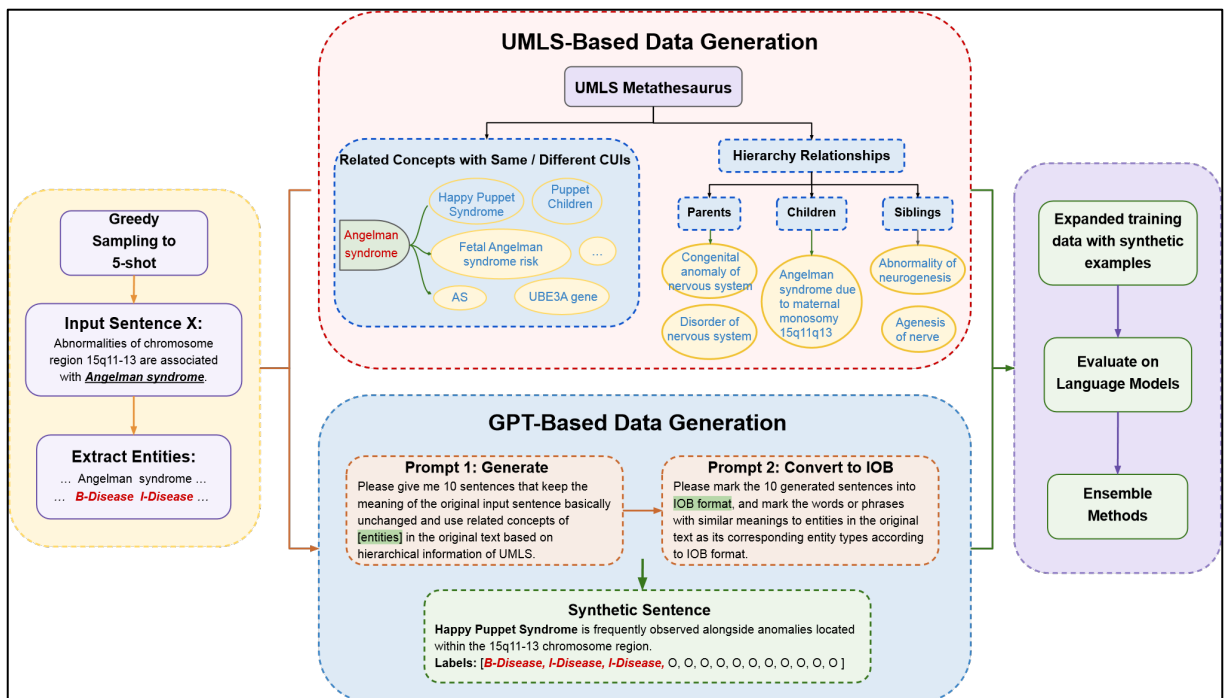

**Figure 2.** The overall architecture of the HILGEN model. After utilizing UMLS-based and GPT-based data generation methods, synthetic sentences are generated from the example sentences. Ensembling is then applied to improve performance.

*Hierarchical Information and Semantic Network in UMLS*

A key feature of UMLS is its hierarchical organization of concepts, representing relationships in a tree-like structure.[37] This hierarchical structure facilitates navigating UMLS and organizes concepts based on their relationships. The hierarchy is based on various types of relationships between concepts, including 'isa' (is a), 'has_parent' and 'has_child' relationships. In addition to the hierarchy, the UMLS also includes a semantic network that describes the relationships between concepts based on their semantic similarity, rather than hierarchy.

*UMLS-Based Data Generation*

When faced with entity types with small numbers of labeled samples, we use the knowledge encoded in the UMLS to expand the training data and add synthetic examples into the training set, thereby expanding the original few-shot training set. Specifically, we incorporate knowledge in multiple layers.

The first layer consists of lexical expressions with the same UMLS concept IDs (concept unique identifiers or CUIs), which are added to create synthetic examples. This layer of augmentation, therefore, introduces potential synonyms of the original named entities in the training data. The second layer of expansion consists of augmenting the training data from the first layer with additional closely related CUIs belonging to the same UMLS semantic type (broad category of concepts, such as pharmacological substances). This layer, thus, adds additional examples that are likely to be conceptually closely related to the entities in the training data, and are therefore likely to occur in similar contexts in medical free text. The third layer of augmentation considers the hierarchical associations in medical concepts. Specifically, we leverage parent-child relationships between concepts, extracting parents, children, and siblings of the given concepts based on the SNOMEDCT_US dictionary (Systematized Nomenclature of Medicine—Clinical Terms), which is a comprehensive clinical terminology that is widely used in the healthcare industry.

*GPT-Based Data Generation*

Our approach to utilizing GPT to generate new examples involves providing the model with complete sentences. As illustrated in Figure 2, we begin with an input sentence, and use a two-step prompt to generate varied, semantically similar sentences that use different expressions for the same entities and convert the sentences into IOB format for subsequent fine-tuning. This strategy enables GPT to leverage contextual information to enhance its comprehension of the given concepts, thereby facilitating the generation of semantically coherent examples. Furthermore, we control the number of generated examples to match the quantity extracted from the UMLS, ensuring that the results are not biased by discrepancies in the amount of training data. In the use of prompts, we adopt a fundamental prompt strategy, which involves providing the sentence itself, indicating its task and the expected output format, while mandating that it generates based on the knowledge from the UMLS. The prompts used to extract hierarchical information and convert generated sentences into IOB format for GPT-based generation in the HILGEN model are listed below.

*Ensemble Method*

To further improve the robustness and accuracy of our biomedical NER models, we employ several ensemble approaches, including weighted voting and intersection. In weighted voting, models are assigned different weights based on their performance, with higher weights given to models with better predictive accuracy, thereby influencing the final prediction more significantly. The intersection method, on the other hand, only accepts predictions that are agreed upon by multiple models, which helps to eliminate noisy predictions. These ensembles combine models trained on synthetic data generated by both UMLS and GPT-3.5. Leveraging the strengths of both data sources, the ensemble model enhances overall performance, reducing the impact of any single model's weaknesses.

*Comparison with ZEROGEN*

We further evaluated the synthetic text generation capabilities of HILGEN by comparing it with ZEROGEN,[35] a zero-shot learning framework that leverages large pre-trained language models (PLMs) to generate synthetic datasets for training smaller task-specific models. We conducted a comparative analysis between our HILGEN approach and ZEROGEN to highlight the effectiveness of our few-shot based methodology against a zero-shot generation strategy. Both methods utilized GPT-3.5 for generating synthetic data to ensure a fair comparison, ensuring consistency in data generation at both the sentence and entity levels.

*Datasets*

We utilized four distinct medical text datasets as benchmarks to evaluate the performance of our models and to support the development of new approaches. These datasets provide a diverse range of

Listing (a): Prompt for extracting related concepts

> Please give me 10 sentences that keep the meaning of the original input sentence basically unchanged and use related concepts of [entities] in the original text based on hierarchical information of UMLS.

Listing (b): Prompt for extracting parents and children

> Based on your knowledge of hierarchical information of UMLS, please find the parents and children of [entities] in the input sentence by using SNOMEDCT_US dictionary. Then, please give me 10 sentences that keep the meaning of the original input sentence basically unchanged and use parents and children of [entities] in the original text.

Listing (c): Prompt for extracting siblings

> Based on your knowledge of hierarchical information of UMLS, please find the siblings of [entities] in the input sentence by using SNOMEDCT_US dictionary. Then, please give me 10 sentences that keep the meaning of the original input sentence basically unchanged and use siblings of [entities] in the original text.

Listing (d): Prompt for converting sentences to IOB format

> Please mark the 10 generated sentences into IOB format, and mark the words or phrases with similar meanings to entities in the original text as its corresponding entity types according to IOB format.

**Figure 3.** Prompts used for synthetic data generation in the HILGEN model.

clinical narratives and biomedical information, allowing for a comprehensive assessment of our methods.

MIMIC III.[1] The MIMIC III dataset is a large, publicly available database with patient data from critical care units, including medications, lab results, clinical notes, diagnostic codes, imaging reports, and survival data. It is widely used for few-shot classification and NER tasks.

BC5CDR.[38] This resource extracts relationships between chemicals and diseases from annotated biomedical articles, aimed at developing systems to automatically identify these interactions for applications like drug discovery, toxicology, and understanding disease mechanisms.

Med-Mentions.[39] Med-Mentions is a large biomedical corpus annotated with UMLS concepts, containing PubMed articles linked to entities like diseases, chemicals, genes, and anatomical terms. It supports tasks such as information extraction, literature mining, and knowledge base construction.

NCBI-Disease.[40] This dataset contains PubMed abstracts annotated with disease names, linked to standardized concepts in MeSH and OMIM databases. It is used to train and evaluate models for recognizing and normalizing disease names in biomedical texts.

**Results**

Table 1 presents relevant statistics for all publicly available datasets we used in this study, including the source and aim of each dataset, training and test set sizes, the number of entity types and the number of entities in each dataset.

**Table 1.** Statistics for the four standardized biomedical datasets we used, including their source, the aim of each task, training and test sizes (in tokens), the number of entity types, and the number of entities in each dataset.

| Datasets | Training Size | Test Size | Entity Types | Entities |
|---|---|---|---|---|
| MIMIC III (information relating to patients) | 36.4k | 6.4k | 12 | 8.7k |
| BC5CDR (extracting relationships between chemicals and diseases) | 228.8k | 122.2k | 2 | 28.8k |
| Med-Mentions (annotated with UMLS concepts) | 847.9k | 593.6k | 1 | 340.9k |
| NCBI Disease (PubMed abstracts annotated with disease names) | 134.0k | 20.5k | 4 | 6.3k |

*Experiment Setup*

We conducted experiments in a few-shot setting, with 5 examples available for each label. We employed GPT-3.5 as the generator to evaluate the generative capabilities of LLMs. We applied two models—BERT and our previously proposed DANN (Data Augmentation with Nearest Neighbor classifier)[41] model—to evaluate the expanded synthetic training data on four biomedical text datasets. We evaluated the performance of our models on four medical datasets: MIMIC III, BC5CDR, NCBI-Disease, and Med-Mentions. The performance metrics used include precision (P), recall (R), and $F_1$ score ($F_1$).

*Experimental Results*

The results in Table 2 clearly demonstrate the effectiveness of HILGEN in generating synthetic data by incorporating prior knowledge through hierarchical information from UMLS and GPT-3.5 to enhance the performance of biomedical NER tasks in few-shot learning settings. Leveraging both UMLS and GPT-3.5 for data generation, we observed significant improvements across all datasets. Incorporating knowledge from related concepts, as well as parent and child relationships from UMLS, often resulted in higher precision and $F_1$ scores, indicating that hierarchical and semantic relationships provide valuable context closely matching the target entities. The performance when using sibling relationships was somewhat mixed, with improvements in certain datasets but not consistently outperforming the other methods. Improvements were particularly noticeable in difficult cases where baseline models struggled to make accurate predictions.

When comparing GPT-3.5 to the incorporation of UMLS, both approaches showed improvements over baseline models. GPT-3.5 generally performed better across most datasets, suggesting its strength in generating diverse, contextually rich examples and understanding complex clinical text. UMLS incorporation shows more consistent improvements across all datasets, as it provides a solid foundation for identifying and categorizing entities based on established medical vocabularies, and the hierarchical information from UMLS contributed to more accurate and contextually relevant synthetic data, highlighting its usefulness in providing structured medical knowledge.

**Table 2.** Comparison of the HILGEN model with DANN for distinct data generation techniques.

| Dataset | | MIMIC III | | | BC5CDR | | | NCBI-Disease | | | Med-Mentions | | |
|---|---|---|---|---|---|---|---|---|---|---|---|---|---|
| | | P | R | $F_1$ | P | R | $F_1$ | P | R | $F_1$ | P | R | $F_1$ |
| **BERT-Large** | | | | | | | | | | | | | |
| Original 5-shot | N/A | 0.74 | 0.08 | 0.14 | 5.14 | 0.37 | 0.69 | 6.57 | 0.76 | 1.36 | 9.34 | 26.88 | 13.86 |
| HILGEN: Generated by hierarchical information from UMLS | with related concepts | 37.33 | 59.65 | 45.92 | 49.69 | 66.42 | 56.85 | 34.86 | 34.56 | **34.71** | 27.62 | 60.13 | 37.85 |
| | with parents and children | 40.41 | 57.05 | 47.30 | 46.01 | 59.93 | 52.05 | 36.39 | 28.14 | 31.74 | 27.83 | 60.56 | 38.14 |
| | with siblings | 36.11 | 56.78 | 44.14 | 48.81 | 55.08 | 51.76 | 37.52 | 29.19 | 32.83 | 26.81 | 59.38 | 36.94 |
| HILGEN: Generated by GPT-3.5 | with related concepts | 38.95 | 60.54 | 47.41 | 51.10 | 52.83 | 51.95 | 30.32 | 32.34 | 31.30 | 27.94 | 60.29 | 38.18 |
| | with parents and children | 41.95 | 62.08 | **50.06** | 46.29 | 62.87 | 53.32 | 28.81 | 30.14 | 29.46 | 28.02 | 62.12 | **38.62** |
| | with siblings | 34.44 | 63.06 | 44.54 | 49.26 | 68.12 | **57.18** | 30.99 | 33.64 | 32.26 | 27.32 | 60.34 | 37.61 |
| HILGEN: Best-Ensemble | N/A | 43.72 | 60.16 | **50.63** | 53.17 | 63.97 | **58.06** | 37.79 | 34.51 | **36.07** | 29.36 | 64.97 | **40.44** |
| **DANN Model** | | | | | | | | | | | | | |
| Original 5-shot | N/A | 19.22 | 21.40 | 19.68 | 27.66 | 50.52 | 35.75 | 18.67 | 27.93 | 22.38 | 48.05 | 57.62 | 52.40 |
| HILGEN: Generated by hierarchical information from UMLS | with related concepts | 52.16 | 58.11 | 54.97 | 52.41 | 73.76 | **61.27** | 33.65 | 46.04 | **38.88** | 60.79 | 67.86 | 64.13 |
| | with parents and children | 51.09 | 56.34 | 53.59 | 51.98 | 72.33 | 60.49 | 35.78 | 35.78 | 35.78 | 60.03 | 67.57 | 63.58 |
| | with siblings | 53.95 | 60.26 | **56.93** | 50.63 | 65.83 | 57.23 | 34.87 | 40.68 | 37.55 | 60.13 | 68.12 | 63.88 |
| HILGEN: Generated by GPT-3.5 | with related concepts | 46.87 | 62.34 | 53.51 | 53.72 | 69.53 | 60.61 | 35.21 | 40.86 | 37.82 | 61.08 | 68.92 | 64.76 |
| | with parents and children | 46.22 | 58.91 | 51.80 | 47.08 | 62.57 | 53.73 | 34.31 | 39.72 | 36.82 | 60.44 | 68.76 | 64.33 |
| | with siblings | 41.54 | 56.64 | 47.94 | 53.11 | 69.30 | 60.13 | 35.24 | 41.01 | 37.91 | 60.28 | 67.81 | 63.83 |
| HILGEN: Best-Ensemble | N/A | 52.79 | 64.60 | **58.68** | 60.52 | 73.85 | **65.09** | 37.10 | 42.99 | **39.83** | 63.49 | 70.28 | **66.72** |

*Comparison with ZEROGEN*

Table 3 provides a detailed comparison of the performance metrics (precision, recall, and $F_1$ score) between the ZEROGEN and HILGEN approaches, which clearly illustrates the superior performance of HILGEN compared to ZEROGEN in all evaluated scenarios. HILGEN achieved up to a 28.19% higher $F_1$ score on the BC5CDR dataset on average, attributed to its use of hierarchical UMLS knowledge, which allowed for more contextually relevant and semantically accurate data generation. In contrast, ZEROGEN's zero-shot approach, though efficient, often generated more generic and less domain-specific data, resulting in lower precision, particularly in datasets like MIMIC III and Med-Mentions. HILGEN's incorporation of UMLS also led to more consistent improvements across datasets, demonstrating its ability to more accurately reflect the complexity and specificity of biomedical language.

*Ensemble Approach*

The results in Table 4 illustrate the impact of the ensemble approach in improving performance by combining the predictions of HILGEN and GPT-3.5, resulting in notable improvements in precision, recall, and $F_1$ scores. For the MIMIC III dataset and BC5CDR dataset, the ensemble method outperforms both the results from GPT-3.5 inference and the use of HILGEN for generating synthetic training data.

HILGEN outperforms GPT-3.5 on the NCBI-Disease and Med-Mentions datasets, achieving an $F_1$ score of 66.72% on Med-Mentions, compared to GPT-3.5's 30.57%. The ensemble method further improves performance, resulting in a more robust and accurate NER system, as reflected by higher $F_1$ scores.

**Table 3.** Comparison of ZEROGEN and HILGEN approaches using BERT-Large and DANN Models on biomedical datasets, demonstrating HILGEN's superior performance across all metrics and datasets.

|  |  | MIMIC III | | | BC5CDR | | | NCBI-Disease | | | Med-Mentions | | |
| --- | --- | --- | --- | --- | --- | --- | --- | --- | --- | --- | --- | --- | --- |
|  |  | P | R | $F_1$ | P | R | $F_1$ | P | R | $F_1$ | P | R | $F_1$ |
| BERT-Large | ZEROGEN | 10.17 | 3.34 | 4.62 | 35.74 | 21.64 | 26.96 | 21.22 | 4.25 | 7.82 | 14.70 | 20.30 | 17.06 |
|  | HILGEN | 43.72 | 60.16 | 50.63 | 53.17 | 63.97 | 58.06 | 37.79 | 34.51 | 36.07 | 29.36 | 64.97 | 40.44 |
| DANN Model | ZEROGEN | 17.32 | 7.80 | 10.75 | 47.95 | 34.05 | 39.82 | 17.13 | 10.78 | 13.24 | 45.42 | 18.46 | 26.25 |
|  | HILGEN | 52.79 | 64.60 | 58.68 | 60.52 | 73.85 | 65.09 | 37.10 | 42.99 | 39.83 | 63.49 | 70.28 | 66.72 |

The table shows precision (P), recall (R), and $F_1$ score ($F_1$) for models trained on synthetic data generated by HILGEN using hierarchical information from UMLS and GPT-3.5. For each dataset, we compare the performance of the original 5-shot model, models using synthetic data generated with related concepts, parent-child relationships, and sibling relationships, and the best ensemble model.

**Table 4.** Enhanced performance of ensemble with predictions from GPT-3.5 on biomedical datasets. Despite HILGEN's competitive results, the ensemble method, which combines HILGEN and GPT-3.5 outputs, improves the overall performance.

|  | MIMIC III | | | BC5CDR | | | NCBI-Disease | | | Med-Mentions | | |
| --- | --- | --- | --- | --- | --- | --- | --- | --- | --- | --- | --- | --- |
|  | P | R | $F_1$ | P | R | $F_1$ | P | R | $F_1$ | P | R | $F_1$ |
| GPT-3.5 | 62.99 | 64.10 | 63.54 | 56.81 | 83.61 | 67.66 | 27.79 | 45.10 | 34.39 | 25.66 | 37.81 | 30.57 |
| HILGEN | 52.79 | 64.60 | 58.68 | 60.52 | 73.85 | 65.09 | 37.10 | 42.99 | 39.83 | 63.49 | 70.28 | 66.72 |
| Ensemble | 59.82 | 70.22 | <u>64.60</u> | 72.03 | 71.65 | <u>71.84</u> | 38.85 | 43.39 | <u>40.99</u> | 53.16 | 65.66 | 58.75 |

**Discussion**

Our findings underscore the value of incorporating structured domain knowledge, such as that found in UMLS, into synthetic data generation. By leveraging hierarchical relationships, HILGEN consistently produced semantically coherent examples, enhancing the performance of NER tasks, particularly in few-shot learning scenarios. The improvements in precision and $F_1$ scores suggest that the hierarchical and semantic relationships embedded in UMLS provide valuable context for identifying and categorizing biomedical entities.

*Challenges of Zero-Shot Data Generation Approaches*

Zero-shot approaches such as ZEROGEN, while eliminating the need for manual annotation, face certain limitations. Firstly, ZEROGEN uses generic prompts with minimal domain-specific constraints, often generating synthetic data that lacks the specific context found in biomedical texts, leading to overly generic or irrelevant content for NER tasks. Secondly, the generated data may exhibit inconsistencies in style and structure with original datasets, failing to capture the language patterns present in actual biomedical texts. Even when the required entity types are provided, ZEROGEN's synthetic datasets often have repetitive sentence structures, failing to capture the linguistic diversity of biomedical texts, which reduces the effectiveness of NER models. Thirdly, generating sentences with multiple entities that resemble original clinical, or biomedical dataset structures is challenging. This is compounded by the fact that, although the entity type distribution may match, the generated text often fails to capture the nuanced context and relationships between entities, leading to a significant drop in model performance.

By incorporating UMLS, HILGEN benefits from a comprehensive and structured medical knowledgebase, ensuring that the generated synthetic examples are semantically coherent and closely aligned with the domain-specific context of biomedical texts. This meticulous approach to maintaining sentence-level and entity-level consistency is crucial, as it allows the synthetic data to accurately reflect the intricate structures and relationships present in the original datasets, thereby improving HILGEN's ability to mimic them and significantly enhancing model performance.

*Impact of Ensemble Learning on Model Generalization*

The ensemble approach, combining models trained on synthetic data generated from both UMLS and GPT-3.5, consistently achieved the highest performance metrics across all datasets. This approach leverages the complementary strengths of UMLS's hierarchical domain-specific knowledge and GPT-3.5's diverse, contextually rich examples.

By integrating the outputs of models trained on different synthetic data sources, the ensemble approach achieved balanced improvements in both precision and recall. Also, it mitigates the issue of data sparsity in few-shot learning scenarios by effectively utilizing the diverse examples generated from UMLS and GPT-3.5. This results in more comprehensive training data, enabling the model to generalize more effectively to unseen instances while maintaining accuracy. Our results highlight the complementary benefits of combining domain-specific knowledge from UMLS with the generative capabilities of LLMs.

*Limitations*

While HILGEN presents a robust approach for generating high-quality synthetic data based on few-shot scenarios for biomedical NER tasks, several limitations must be acknowledged. First, the scope of our current data generation and expansion is somewhat limited. Specifically, we identified and used only the top 10 related concepts for each entity, and our expansion and generation process relied on a 5-shot setting. It is plausible that utilizing a higher number of annotated examples, such as 10 or 20, and incorporating a wider array of related concepts could potentially yield superior results. Our primary objective in this article was to establish the feasibility and effectiveness of the HILGEN approach. We hypothesize that further expanding the synthetic dataset would result in improved model performance. Nonetheless, this expansion would also entail additional computational and resource costs. The final limitation is our focus on methodology over prompt engineering. The prompts we used were relatively basic. It is plausible that more sophisticated prompt engineering could lead to better results. Future research can explore advanced prompt engineering techniques to fully leverage the capabilities of LLMs.

**Conclusion**

Our experiments demonstrate that the HILGEN model, by combining synthetic data from UMLS and LLMs, can improve NER performance in few-shot settings. Our findings challenge the belief that the rise of LLMs diminishes the value of expert-curated knowledge, which remains essential for improving predictions on unseen data. The GPT models add context-aware understanding of entities, enabling richer entity recognition and expanding the training data without requiring additional manual annotations, particularly benefiting rare or complex cases. Using information from the hierarchical structure of UMLS and LLMs as external knowledge bases can generate high-quality synthetic datasets to address key challenges in few-shot learning with medical text datasets, including limited training data and the need for domain-specific knowledge. Our approach can be extended to few-shot learning involving diverse biomedical datasets and problems.

**References**
1. Johnson, A. E. *et al.* MIMIC-III, a freely accessible critical care database. *Scientific data* **3**, 1–9 (2016).
2. Murdoch, T. B. & Detsky, A. S. The inevitable application of big data to health care. *Jama* **309**, 1351–1352 (2013).
3. Dong, N. & Xing, E. P. Few-shot semantic segmentation with prototype learning. in *BMVC* vol. 3 4 (2018).
4. Ge, Y., Guo, Y., Das, S., Al-Garadi, M. A. & Sarker, A. Few-shot learning for medical text: A review of advances, trends, and opportunities. *Journal of Biomedical Informatics* 104458 (2023).
5. Lake, B. M., Salakhutdinov, R. R. & Tenenbaum, J. One-shot learning by inverting a compositional causal process. *Advances in neural information processing systems* **26**, (2013).
6. Sung, F. *et al.* Learning to compare: Relation network for few-shot learning. in *Proceedings of the IEEE conference on computer vision and pattern recognition* 1199–1208 (2018).
7. Yoo, T. K., Choi, J. Y. & Kim, H. K. Feasibility study to improve deep learning in OCT diagnosis of rare retinal diseases with few-shot classification. *Med Biol Eng Comput* **59**, 401–415 (2021).
8. Zhao, Z., Wallace, E., Feng, S., Klein, D. & Singh, S. Calibrate before use: Improving few-shot performance of language models. in *International conference on machine learning* 12697–12706 (PMLR, 2021).
9. Singhal, K. *et al.* Large language models encode clinical knowledge. *Nature* **620**, 172–180 (2023).
10. Azamfirei, R., Kudchadkar, S. R. & Fackler, J. Large language models and the perils of their hallucinations. *Crit Care* **27**, 120 (2023).
11. Anderson, B. R., Shah, J. H. & Kreminski, M. Homogenization Effects of Large Language Models on Human Creative Ideation. in *Creativity and Cognition* 413–425 (ACM, Chicago IL USA, 2024). doi:10.1145/3635636.3656204.
12. Snell, J., Swersky, K. & Zemel, R. Prototypical networks for few-shot learning. *Advances in neural information processing systems* **30**, (2017).

13. Wang, Y., Yao, Q., Kwok, J. T. & Ni, L. M. Generalizing from a Few Examples: A Survey on Few-shot Learning. *ACM Comput. Surv.* **53**, 1–34 (2021).
14. Schmidt, H. K., Rothgangel, M. & Grube, D. Prior knowledge in recalling arguments in bioethical dilemmas. *Frontiers in psychology* **6**, 1292 (2015).
15. Hospedales, T., Antoniou, A., Micaelli, P. & Storkey, A. Meta-learning in neural networks: A survey. *IEEE transactions on pattern analysis and machine intelligence* **44**, 5149–5169 (2021).
16. Vinyals, O., Blundell, C., Lillicrap, T. & Wierstra, D. Matching networks for one shot learning. *Advances in neural information processing systems* **29**, (2016).
17. Pan, S. J. & Yang, Q. A survey on transfer learning. *IEEE Transactions on knowledge and data engineering* **22**, 1345–1359 (2009).
18. Prato, G., Charlaix, E. & Rezagholizadeh, M. Fully Quantized Transformer for Machine Translation. in *Findings of the Association for Computational Linguistics: EMNLP 2020* 1–14 (2020).
19. Liu, J. *et al.* Unified instance and knowledge alignment pretraining for aspect-based sentiment analysis. *IEEE/ACM transactions on audio, speech, and language processing* **31**, 2629–2642 (2023).
20. Li, X. L. & Liang, P. Prefix-Tuning: Optimizing Continuous Prompts for Generation. Preprint at https://doi.org/10.48550/arXiv.2101.00190 (2021).
21. Gu, Y., Han, X., Liu, Z. & Huang, M. PPT: Pre-trained Prompt Tuning for Few-shot Learning. in *Proceedings of the 60th Annual Meeting of the Association for Computational Linguistics (Volume 1: Long Papers)* 8410–8423 (2022).
22. Michalopoulos, G., Wang, Y., Kaka, H., Chen, H. & Wong, A. UmlsBERT: Clinical Domain Knowledge Augmentation of Contextual Embeddings Using the Unified Medical Language System Metathesaurus. in *Proceedings of the 2021 Conference of the North American Chapter of the Association for Computational Linguistics: Human Language Technologies* 1744–1753 (2021).
23. Rumshisky, A., Roberts, K., Bethard, S. & Naumann, T. Proceedings of the 3rd Clinical Natural Language Processing Workshop. in *Proceedings of the 3rd Clinical Natural Language Processing Workshop* (2020).
24. McCray, A. T., Bodenreider, O., Malley, J. D. & Browne, A. C. Evaluating UMLS strings for natural language processing. *Proc AMIA Symp* 448–452 (2001).
25. Brown, T. *et al.* Language Models are Few-Shot Learners. in *Advances in Neural Information Processing Systems* vol. 33 1877–1901 (Curran Associates, Inc., 2020).
26. OpenAI *et al.* GPT-4 Technical Report. Preprint at http://arxiv.org/abs/2303.08774 (2024).
27. Thoppilan, R. *et al.* LaMDA: Language Models for Dialog Applications. Preprint at https://doi.org/10.48550/arXiv.2201.08239 (2022).
28. Touvron, H. *et al.* LLaMA: Open and Efficient Foundation Language Models. Preprint at https://doi.org/10.48550/arXiv.2302.13971 (2023).
29. Ji, Z. *et al.* Survey of Hallucination in Natural Language Generation. *ACM Comput. Surv.* **55**, 1–38 (2023).
30. Gligic, L., Kormilitzin, A., Goldberg, P. & Nevado-Holgado, A. Named entity recognition in electronic health records using transfer learning bootstrapped Neural Networks. *Neural Networks* **121**, 132–139 (2020).
31. Gupta, P., Malhotra, P., Narwariya, J., Vig, L. & Shroff, G. Transfer Learning for Clinical Time Series Analysis Using Deep Neural Networks. *J Healthc Inform Res* **4**, 112–137 (2020).
32. Chintagunta, B., Katariya, N., Amatriain, X. & Kannan, A. Medically Aware GPT-3 as a Data Generator for Medical Dialogue Summarization. in *Proceedings of the 6th Machine Learning for Healthcare Conference* 354–372 (PMLR, 2021).
33. Anaby-Tavor, A. *et al.* Do Not Have Enough Data? Deep Learning to the Rescue! *Proceedings of the AAAI Conference on Artificial Intelligence* **34**, 7383–7390 (2020).
34. Yang, Y. *et al.* Generative Data Augmentation for Commonsense Reasoning. in *Findings of the Association for Computational Linguistics: EMNLP 2020* (eds. Cohn, T., He, Y. & Liu, Y.) 1008–1025 (Association for Computational Linguistics, Online, 2020). doi:10.18653/v1/2020.findings-emnlp.90.
35. Ye, J. *et al.* ZeroGen: Efficient Zero-shot Learning via Dataset Generation. in *Proceedings of the 2022 Conference on Empirical Methods in Natural Language Processing* (eds. Goldberg, Y., Kozareva, Z. & Zhang, Y.) 11653–11669 (Association for Computational Linguistics, Abu Dhabi, United Arab Emirates, 2022). doi:10.18653/v1/2022.emnlp-main.801.
36. Bodenreider, O. The Unified Medical Language System (UMLS): integrating biomedical terminology. *Nucleic Acids Research* **32**, D267–D270 (2004).
37. Mishra, A. *et al.* A Generative Approach to Zero-Shot and Few-Shot Action Recognition. in *2018 IEEE Winter Conference on Applications of Computer Vision (WACV)* 372–380 (2018). doi:10.1109/WACV.2018.00047.
38. Li, J. *et al.* BioCreative V CDR task corpus: a resource for chemical disease relation extraction. *Database* **2016**, baw068 (2016).
39. Mohan, S. & Li, D. MedMentions: A Large Biomedical Corpus Annotated with UMLS Concepts. Preprint at https://doi.org/10.48550/arXiv.1902.09476 (2019).
40. Doğan, R. I., Leaman, R. & Lu, Z. NCBI disease corpus: A resource for disease name recognition and concept normalization. *Journal of Biomedical Informatics* **47**, 1–10 (2014).
41. Ge, Y., Al-Garadi, M. A. & Sarker, A. Data Augmentation with Nearest Neighbor Classifier for Few-Shot Named Entity Recognition. *Stud Health Technol Inform* **310**, 690–694 (2024).